\documentclass[11pt]{article} 
\usepackage{rldmsubmit,palatino}
\usepackage{graphicx}

\title{Towards neoRL networks; the emergence of purposive graphs.}
\author{Per R. Leikanger  \\
    \mbox{}\\
    UiT -- Norges Artigste Universitet\\
    \texttt{Per.Leikanger@uit.no}} 

\usepackage{amssymb} 
\usepackage{amsmath} 
\usepackage{xcolor}

\usepackage[export]{adjustbox}  

\usepackage{caption}
\usepackage{subcaption}
\usepackage{wrapfig}
\usepackage{tikz}

\begin{document}

\maketitle

\begin{abstract}

    The neoRL framework for purposive AI implements latent learning by emulated cognitive maps, with general value functions (GVF) expressing operant desires toward separate states.
    The agent's expectancy of reward, expressed as learned projections in the considered space, allows the neoRL agent to extract purposive behavior from the learned map according to the reward hypothesis. 
    We explore this allegory further, considering neoRL modules as nodes in a network with desire as input and state-action Q-value as output;
        we see that action sets with Euclidean significance imply an interpretation of state-action vectors as Euclidean projections of desire.
    \emph{Autonomous desire} from neoRL nodes within the agent allows for deeper neoRL behavioral graphs.
    Experiments confirm the effect of neoRL networks governed by autonomous desire, verifying the four principles for purposive networks. 
    A neoRL agent governed by purposive networks can navigate Euclidean spaces in real-time while learning,
        exemplifying how modern AI still can profit from inspiration from early psychology.
\end{abstract}



\startmain 


\section{Behavioristic AI by neoRL nodes}

    Thorndike's \emph{law-an-effect} in functionalist psychology have been reported as an important inspiration for Reinforcement Learning (RL) in AI\ \cite{sutton2018reinforcement}.
    Thorndike considered the reinforcement of randomly encountered reflexes as a plausible explanation for simple behavior and acquired reflexes. 
    The law-of-effect represents an essential first step toward 
                \begin{wrapfigure}{r}{0.45\textwidth} 
                    \centering
                    \includegraphics[width=0.35\columnwidth]{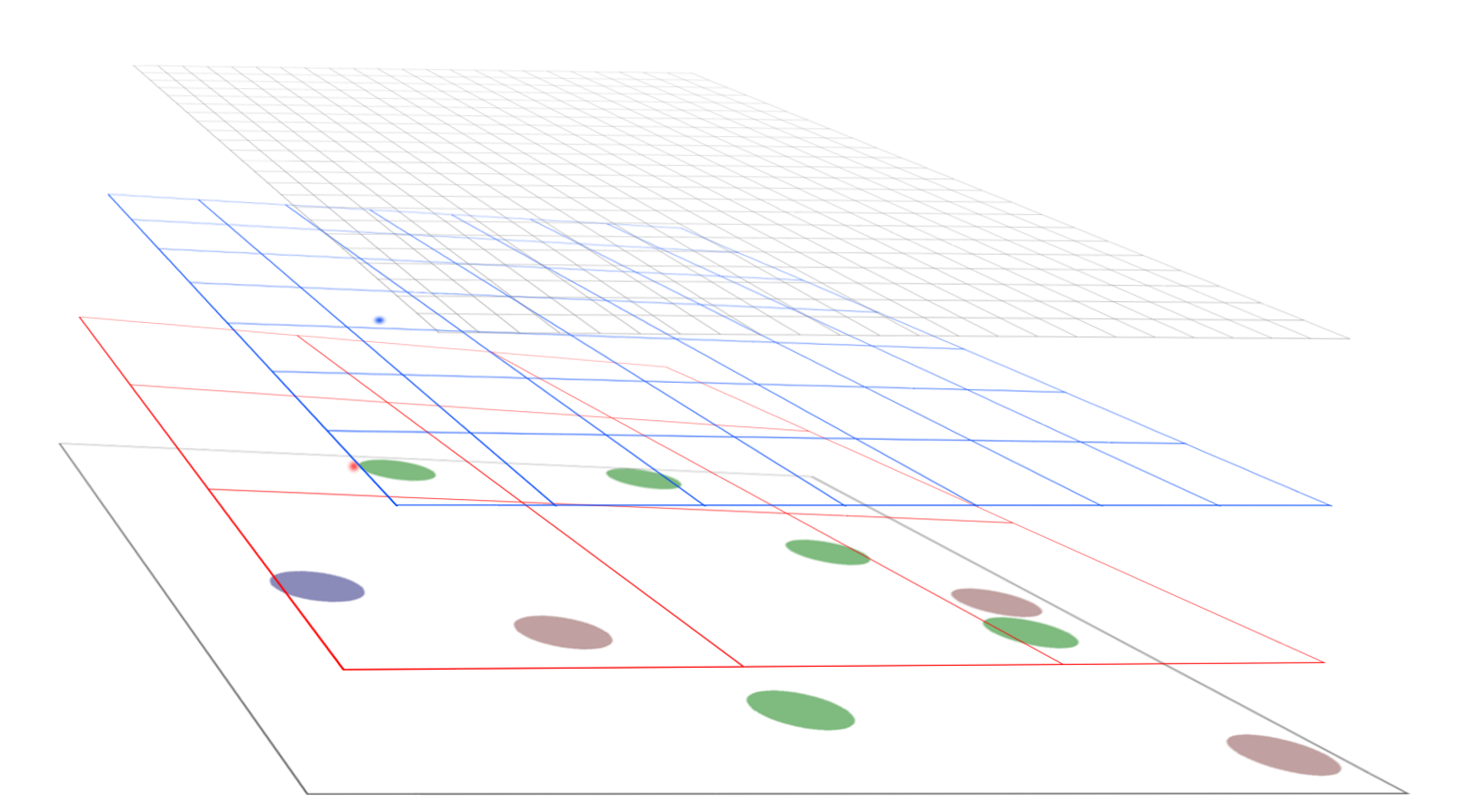}
                    \caption{
                                \textbf{The first multi-map neoRL agent ;}
                                The neoRL agent is capable of expressing latent learning across several representations of the same Euclidean space, forming agent value function as a weighted sum of operant value from all NRES maps. 
                                \mbox{(figure from \cite{leikanger_CCN_2019})}
                            }
                    \label{fig:illustration_multishepherd_3_7_23}
                \end{wrapfigure}                        
            the study of behavior becoming a natural science, quickly replaced by \emph{behaviorism} for explaining advanced policies and human behavior.
    Edward C. Tolman (1866-1959) further proposed that behavior is separate from learning, attempting to explain observations where an animal could express different behavior as a function of varying motivation \cite{tolman1930degrees}.
    Combining Tolman's \emph{latent learning} with \emph{operant conditioning} from E. C. Skinner, considering policies as being \emph{operant} toward an objective, the \emph{neoRL} framework allows for purposive behaviorism for Euclidean navigation\ \cite{leikanger_CCN_2019}.
    By expressing latent learning as a set of operant reflexes in the environment, 
        i.e., with a set of general value functions (GVF)\ \cite{sutton2011horde} trained by mutually exclusive conditionals as reward signals, 
        agent purpose becomes an expression of the parameter configurations where operant value functions are extracted. 
    Considering a set of conditionals inspired by the 2014 Nobel Price in neuroscience, the discovery of place cells and other mechanisms behind state representation for neural navigation, Leikanger (2019) demonstrated how operant \mbox{neoRL} sub-agents could apply for autonomous navigation. 
    Latent learning by a set of operant GVFs combined with elements-of-interest, projections associated with reward in the considered Euclidean space, allows for autonomous navigation governed by the purpose of attaining reward.
    See my thesis \cite{leikanger_thesis_preprint} for more on the theory behind NRES-oriented RL \mbox{(neoRL)} agents, scheduled to be presented in a public online\footnote{Information about the streamed PhD defence will be posted on \emph{www.neoRL.net}} PhD defence only days before RLDM. 


    A neoRL learning module can be considered as a behavioral node in a purposive network; early results for neoRL navigation explored the effect of considering multiple state spaces in parallel for the neoRL agent.
    In some ways analogous to the Hybrid Reward Architecture\ \cite{van2017hybrid}, the neoRL navigation agent combines several learners that establish GVFs toward separate concerns\ \cite{leikanger_alife_2021}. 
    From applying the superposition principle in the value domain, the neoRL agent is capable of combining value function from many learners in one state \mbox{space\ \cite{leikanger_CCN_2019}}, across multiple representations of the same state space\ \cite{leikanger_alife_2021}, 
        or across information represented in orthogonal Euclidean spaces -- thus capable of fully decomposing the state space to simpler considerations\ \cite{leikanger_AGI_2021}.
    The behavioral node consists of three parts;
      first, the latently learned cognitive map formed by GVF on operant desires toward NRES cells.
    Operant desires are trained by off-policy GVF, expressing latent learning on how to accomplish different conditionals in this environment representation.
      Second, the neoRL agent is governed by purpose -- mental projections of parameter configurations associated with reward are expressed as elements-of-interest in the NRES map. 
        \begin{wrapfigure}{l}{0.45\textwidth}   
            \centering
            \includegraphics[width=0.35\columnwidth]{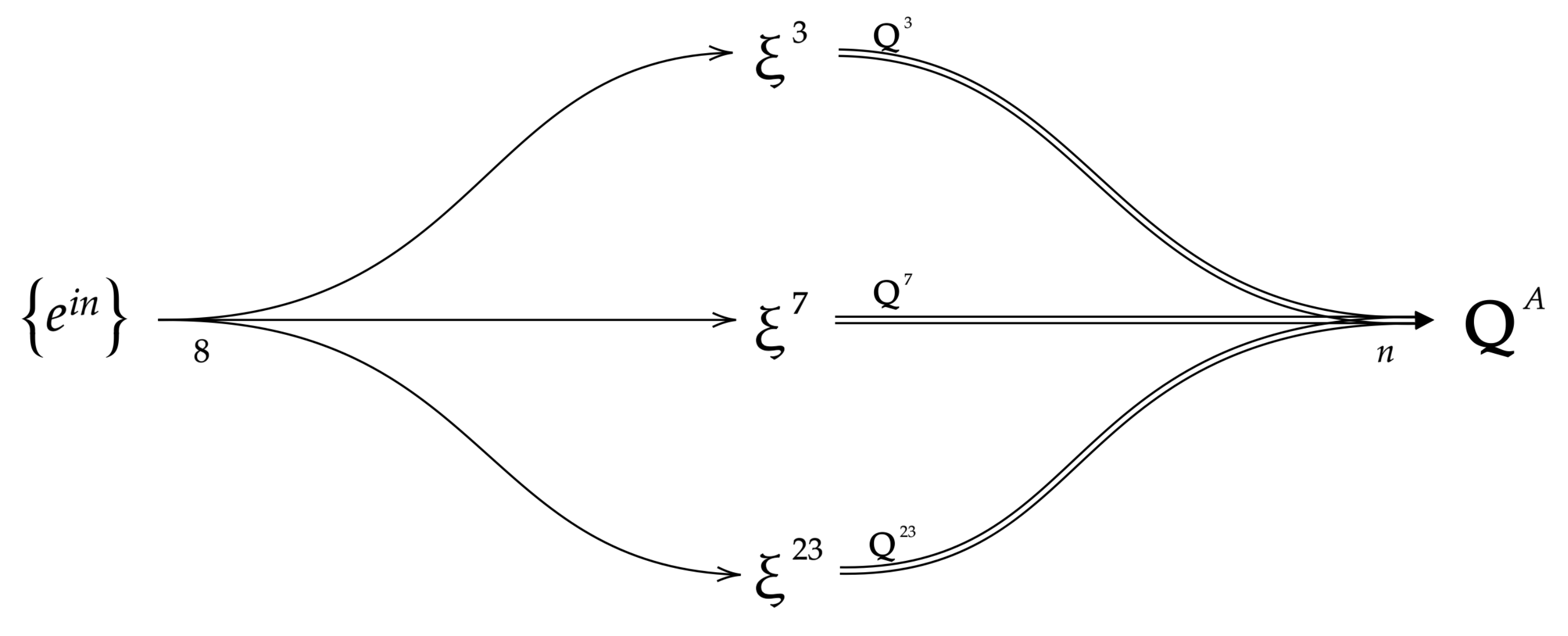}
            \caption{
                        \textbf{A schematic representation of fig. \ref{fig:illustration_multishepherd_3_7_23},}  the neoRL agent from\ \cite{leikanger_alife_2021}.
                        Agent value function is formed from the combined value function $Q^N$ from each of the three $\xi^N$ neoRL nodes, where $N\in \{3, 7, 23\}$\ \cite{leikanger_alife_2021}.
                    }
            \label{fig:illustration_schema_3_7_23_joint_Q_function}
        \end{wrapfigure}                        
    These free-ranging representations of desire in the considered space are mapped to NRES nodes, activating GVFs corresponding to the associated valence -- the element's expectancy of reward upon achievement.
    Note that the same neoRL node, containing latent knowledge in one NRES representation, can be harvested by different sets of elements-of-interest. 
      Third, the value function can be extracted by elements-of-interest from the digital analogy to Tolman's cognitive map -- resulting in an actionable Q-vector output of the neoRL node.
    When mutually exclusive NRES receptive fields are used for latent learning, the GVF components become operant toward that NRES cell -- Operant Value-function Components (OVC).
    The singular (orthogonal) value component can be combined with others to form the full value function of the agent, 
        further implying that multiple modalities can be learned in parallel and combined to one agent value function\ \cite{leikanger_AGI_2021}.
    Figure \ref{fig:illustration_schema_3_7_23_joint_Q_function} shows the aggregation of the value function for the neoRL agent in \cite{leikanger_alife_2021}.
    The location and valence of elements-of-interest can be considered as inputs to the neoRL behavioral node, and the superposition of weighted OVC establishes an output of the neoRL node.
    %

\begin{wrapfigure}{r}{0.45\textwidth}   
            \centering
            \includegraphics[width=0.35\columnwidth]{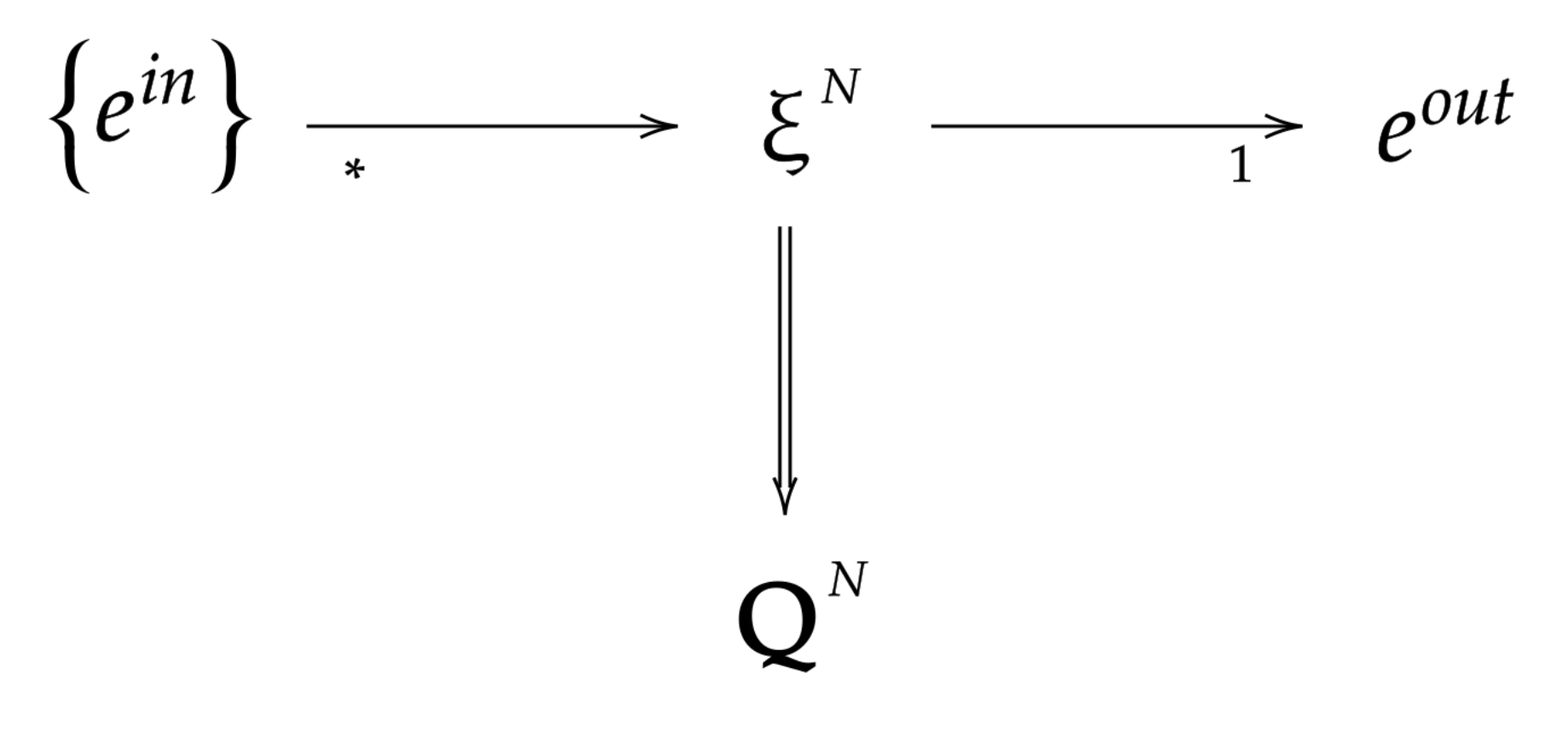}
            \caption{
                        \textbf{A neoRL learning module with one input and two output;}
                            actions with a Euclidean significance implies state-action vectors to be representable in the same Euclidean space; $e^{out}$ can be used as input for compatible neoRL nodes.
                    }
            \label{fig:neoRL_module}

        \end{wrapfigure}

    The purposive AI expressed by the neoRL agent in figure \ref{fig:illustration_schema_3_7_23_joint_Q_function} can well be considered as a single-layered behavioral network with a single output state-action value function.
    Figure \ref{fig:illustration_schema_3_7_23_joint_Q_function} presents a functional representation of the multi-map navigational agent from figure \ref{fig:illustration_multishepherd_3_7_23}.
    The individual neoRL sub-nodes  $\xi^3, \xi^7, \xi^{23}$ are trained by latent learning, and purposive state-action value functions $\{Q^3, Q^7, Q^{23}\}$ can be extracted for the digital analogy to a cognitive map by purposive elements $\{e^{in}\}$.
    The network would have an input, the full set of elements of interest $\{e^{in}\}$;
        the purposive network would have a latent state, formed by latent learning expressed as off-policy operant desires;
        the neoRL network would have an actionable state-action Q-vector as output.
    When further assuming a Euclidean significance for the action set, as with \mbox{$\mathbb{A} = \{N, S, E, W\}$} in\ \cite{leikanger_alife_2021}, the state-action value could be interpreted a Euclidean \emph{desire-vector};
    $$ 
        \vec{d} = \vec{\sum} \mathbf{Q}^{in} \quad,
    $$ 
    where $\vec{\sum}$ represents the vector sum.
        The valence of the desire vector $\vec{d}$ should express the combined valence across $\{e^{in}\}$.
    \begin{align*}
        \vec{e^{out}} &= \vec{d} \\
        e^{out}_\psi &= \sum e^{in}_\psi    \;,
    \end{align*}
    where $\vec{e^i}$ signifies the coordinate and $e^i_\psi$ represents the valence of purposive element $i$.
    A functional schematic of the neoRL module is illustrated in figure\ \ref{fig:neoRL_module};
        a single output-desire can be formed from any number of input elements $\{e^{in}\}$.
    Different sets of purposive elements-of-interest $\{e^{in}\}$ can establish different output desire $e^{out}$ and actionable state-action values $Q^N$ from the same cognitive map.
    Earlier networks of the neoRL node could be seen as a behavioral analog to a one-layered perceptron\ \cite{rosenblatt1957perceptron}.
    The enclosed theory allows for multi-layered neoRL networks governed by autonomous desire.


\section{Experiments}

    A comprehensive environment for research on autonomous navigation is the PLE implementation\ \cite{webWaterWorld_in_PLE} of Karpathy's WaterWorld environment;
    An agent controls acceleration of the self in the four Cardinal directions \mbox{$[N, S, E, W]$}, i.e., \mbox{[up, down, right, left]}.
    Three objects move around in the Euclidean plane according to predefined mechanics.
    Encountering an object replaces it with a new object with a random color, location, and speed vector.
    Green objects are desirable with a positive reward $\mathbb{R} = +1.0$, and red objects are repulsive with a negative valence of $-1.0$.
    Capturing the last green object resets all remaining (red) objects by the same reset mechanisms.   
    The only reward comes from encountering objects, making the accumulation of $\mathbb{R}$ an objective measure for navigational capabilities and its time course an indication of real-time learning capabilities. 


    Rewards are only received sparsely, with discrete $+1.0$ or $-1.0$ steps after encountering objects in the WaterWorld environment; measuring the immediate proficiency of the agent by accumulated $\mathbb{R}$ can be a challenge.
    Capturing the transient time course of agent skill can be done by averaging independent runts; the enclosed experiments average $100$ separate runs to measure transient navigational proficiency, i.e., across $100$ separate agents.
    No pre-training or other precursors are available for the agents, making all navigation happen live as the agent gathers experience in the environment for the first time. 
    Curves are presented with minutes along the x-axis, signifying the wall-clock time since the beginning of each run. 

\begin{figure}[htb!p]
        \centering
        \begin{subfigure}[b]{0.49\textwidth}
            \centering
            \includegraphics[width=\textwidth]{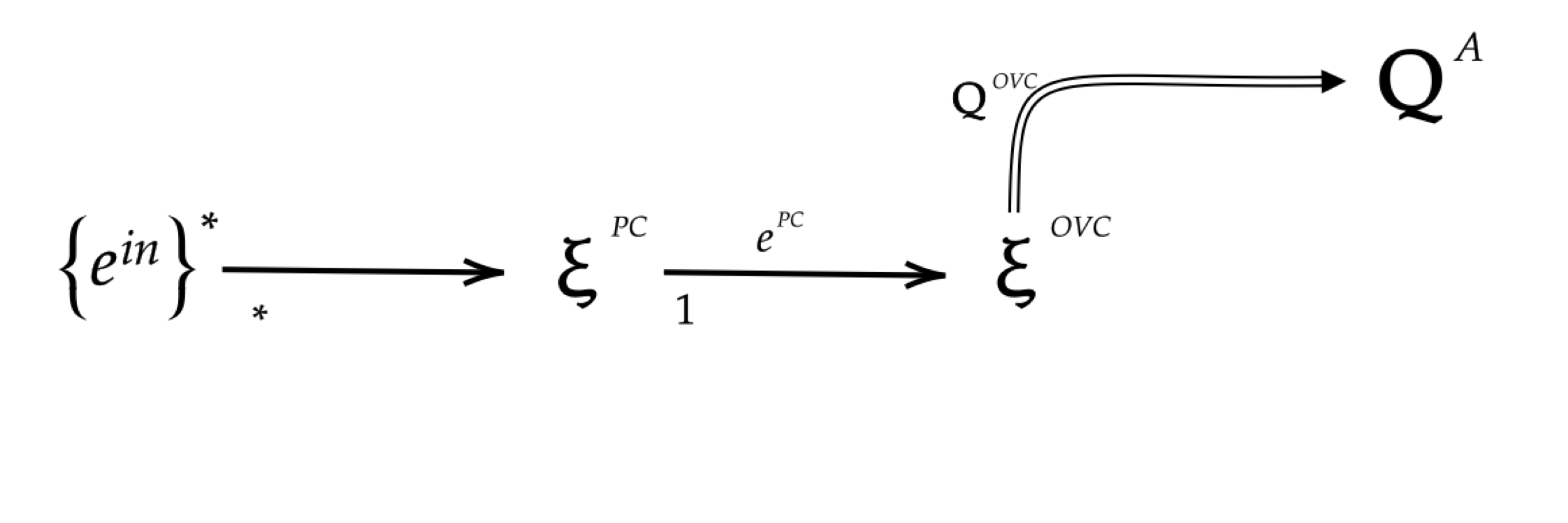}
            \caption[A]{Aspect 1: a single desire vector $e^{PC}$ as input to $\xi^{OVC}$.}
            \label{fig:experiment1}
        \end{subfigure}
        \hfill
        \begin{subfigure}[b]{0.49\textwidth}
            \centering
            \includegraphics[width=\textwidth]{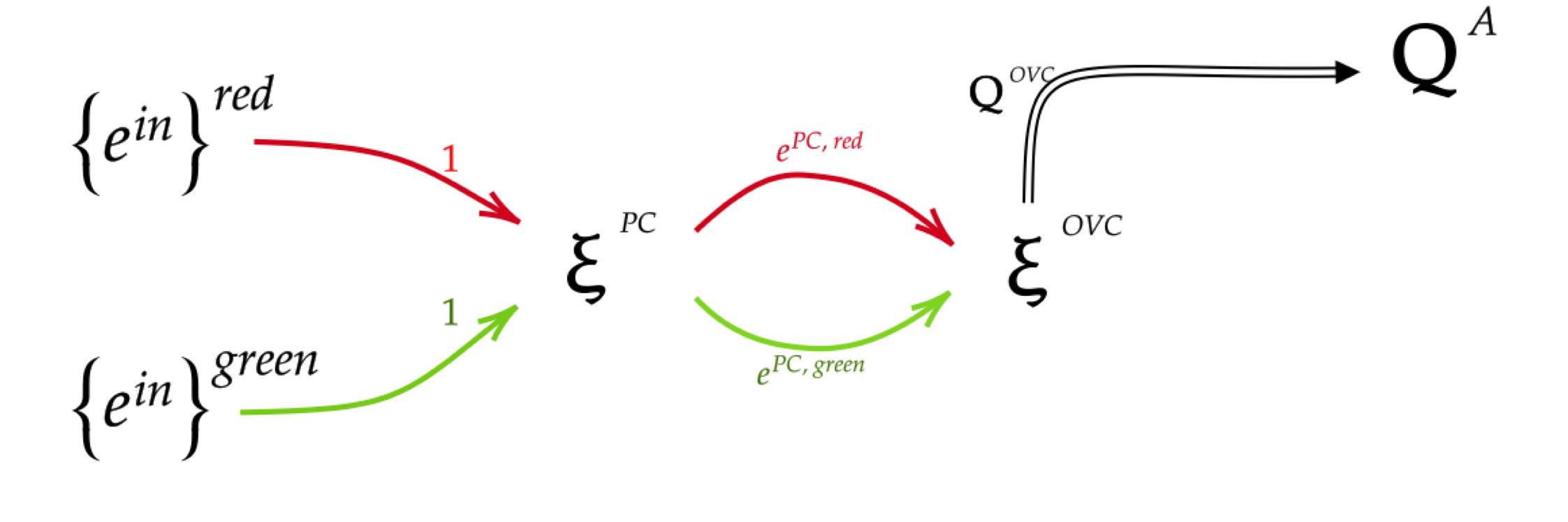}
            \caption{Aspect 2: separate desire extraction; $e^{PC_{red}}$ and $e^{PC_{green}}$.}
            \label{fig:experiment2}
        \end{subfigure}
        \hfill
        \begin{subfigure}[b]{0.49\textwidth}
            \centering
            \includegraphics[width=\textwidth]{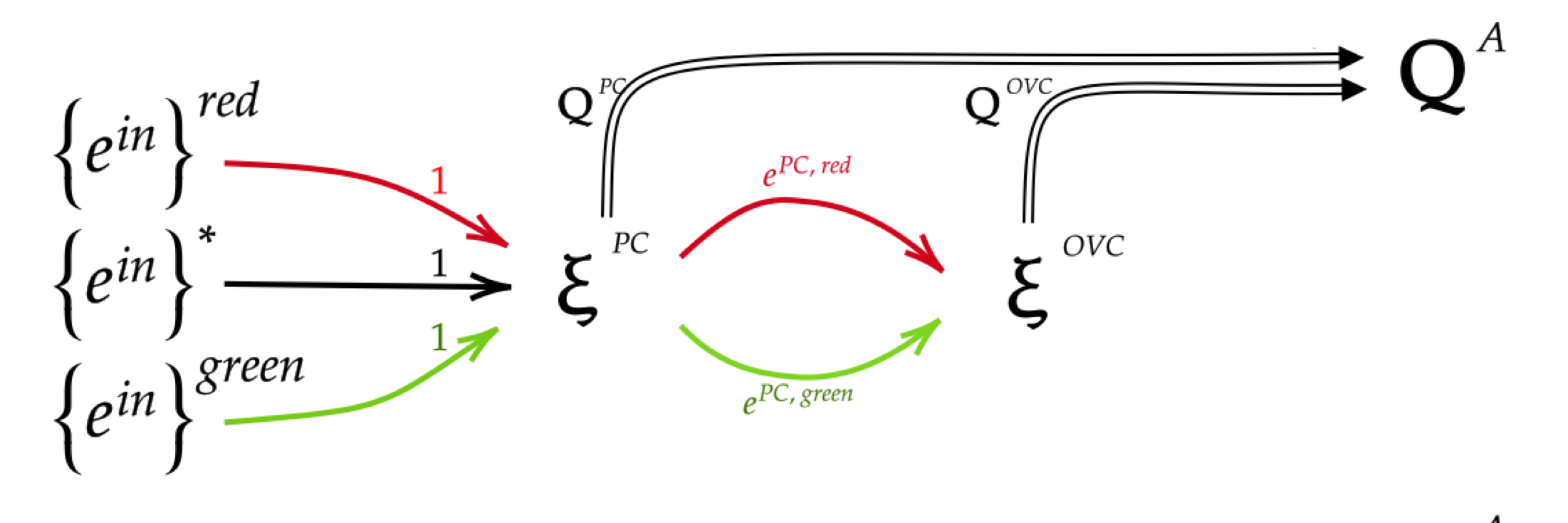}
            \caption{Aspect 3: joint value function from sequential neoRL nodes.}
            \label{fig:experiment3}
        \end{subfigure}
        \hfill
        \begin{subfigure}[b]{0.49\textwidth}
            \centering
            \includegraphics[width=\textwidth]{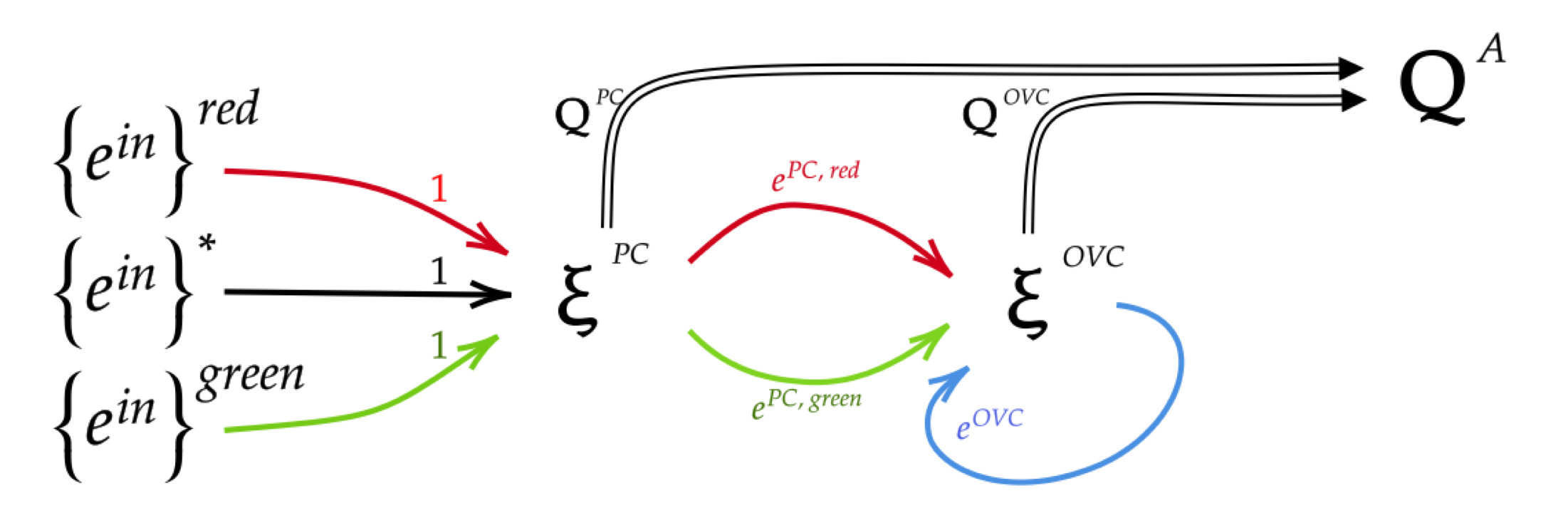}
            \caption{Aspect 4: recursive desires for neoRL autonomy.}
            \label{fig:experiment4}
        \end{subfigure}
        \hfill
        \vspace{60pt}
        \begin{subfigure}[b]{1.0\textwidth}
            \centering
            \includegraphics[width=\textwidth]{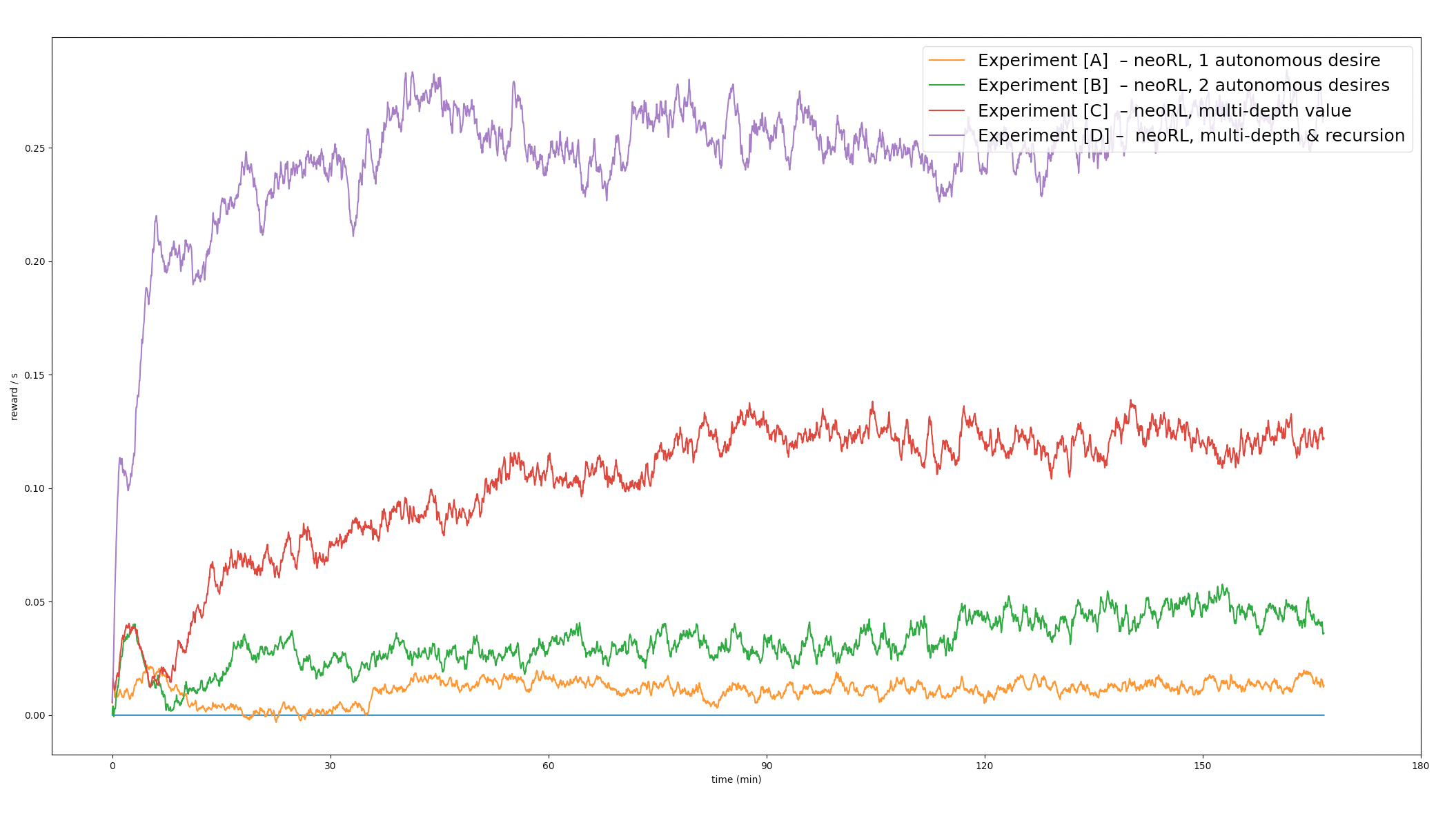}
            \caption{Transient proficiency of neoRL agent A-D.}
            \label{fig:results}
        \end{subfigure}

        \caption{ 
            [\textbf{[Top]}] Illustrations of the neoRL architecture tested in experiment A-D.
            [\textbf{a}] A first attempt on desire from experience; neoRL node $\xi^{PC}$ forms a single desire $e^{{PC}}$ for value-generating neoRL node $\xi^{OVC}$. 
            [\textbf{b}] Latent knowledge can be extracted separately for separate classes for desire; experiment B forms two desire-vectors $e^{PC_{red}}$ and $e^{PC_{green}}$ from $\xi^{OVC}$ -- grouping according to valence. 
            [\textbf{c}] The value function output from neoRL node $\xi^{PC}$ and node $\xi^{OVC}$ contribute equally to agent value function.
            [\textbf{d}] Recursive desires are possible for neoRL nodes: the $\xi^{OVC}$ is governed by three elements-of-interest, $e^{PC_{red}}$, $e^{PC_{green}}$, and recurrent desire $e^{OVC}$.
            [\textbf{[Down]}]    Results from the four experiments: 
            [\textbf{e}] Purposive neoRL networks allows for purposive autonomy by deep and/or recurrent desires.
        }
        \label{fig:illustration_of_experimental_agents}
    \end{figure}

        
    We shall explore four aspects in the WaterWorld environment;
        first, we challenge the basic principle of propagating purpose by the layout illustrated in figure \ref{fig:illustration_of_experimental_agents}\textbf{a}.
    The first neoRL node, $\xi^{PC}$, forms a purposive desire based on all reported objects from WaterWorld; a single desire vector $\vec{d}$ with accompanying valence $e_\psi$ propagates to the compatible neoRL node $\xi^{OVC}$ as autonomous desire $e^{PC}$.
    \emph{Experiment [\textbf{b}]} explores the effect of extracting separate desires from the same learned cognitive map. 
    A purposive desire vector from the neoRL node comes from extracting latent knowledge from considered coordinates; 
        the agent extracts two purposive vectors from $\xi^{PC}$, one from desirable objects $e^{in_{green}}$ and a separate from aversive objects $e^{in_{red}}$. 
    The output from multiple neoRL nodes can be combined for the policy-forming value function in the neoRL framework; 
        \emph{experiment [\textbf{c}]} explores the effect of aggregating value function from multiple depths of the neoRL net.
    The output desire from a neoRL node $\xi^N$ can be applied to any compatible neoRL node $\xi^M$, including itself: 
        \emph{experiment [\textbf{d}]} explores recurrent connections for neoRL nodes.
    Collaborative experience is explored with and without recursive desires, as illustrated in figure \ref{fig:experiment3} and \ref{fig:experiment4}.
    All results are reported in figure \ref{fig:results}.

        


\newpage 
\section{Discussion}
    The neoRL agent navigates continuous space by projections of desire, vectors of purpose associated with an agent's expectancy or reward. 
    When actions have a Euclidean significance, purposive Q-vectors can form autonomous projections of desire, experience-based inferences that can establish input to deeper neoRL nodes.
    Experiments demonstrate how deeper or recurrent desires are crucial for navigational proficiency, 
        suggesting purposive neoRL networks as a plausible approach to autonomous navigation.


    This work explores four principles of purposive neoRL networks.
    \emph{First}, {experience-based desire vectors can shape purposive navigation in Euclidean space.}
    The neoRL network from illustration \ref{fig:experiment1} improves navigational proficiency over time; however, considering all objectives under one, i.e., forming a single desire vector based on all red \emph{and} green objects, becomes too simple for proficient navigation. 
    \emph{Second}, {a single neoRL node can generate different $e^{out}$ desires by considering different sets of objectives $\{e^{in}\}$.} 
    Experiment \textbf{b} demonstrates the effect of separating desires according to valence, resulting in the increased performance by the neoRL navigational agent. 
    The simplicity and clarity expressed by separation of desires, as illustrated in figure \ref{fig:experiment2}, facilitates explainability of the trained solution 
        -- a crucial element if traditional AI is to be applied for desire classification.
    \emph{Third}, {the neoRL agent can base agent value function on any neoRL node in the network.}
    Agents extracting purpose from multiple depths of the neoRL network, as illustrated in \ref{fig:experiment3}, becomes better navigators than more superficial agents.
    \emph{Fourth}, {desire vectors $e^{out}$ from one neoRL node can form objectives for any compatible neoRL node -- including itself.}
    Experiment [d] explores recursive desires, where the output of $\xi^{OVC}$ -- desire vector $e^{OVC}$ -- establish an additional purpose for neoRL node $\xi^{OVC}$.
    The proficiency of the recursive agent from \ref{fig:experiment4} is reported as curve [D], showing how recursive desires drastically improve the agent's navigational proficiency.
    Results can be examined in figure \ref{fig:results} and in real-time video demonstrations at \emph{www.neoRL.net} .

    Note that no comparison has been made with alternative approaches for control;
        this work is only concerned with uncovering the basic principles of purposive neoRL nets for behavioral AI.
    Still, any attempt on finding RL or AI solutions capable of allocentric Euclidean navigation in real-time has failed.
    Further work could involve finding and comparing alternative approaches for real-time autonomous navigation in the WaterWorld environment.
    Likewise, this work involves no search for optimal parameters for neoRL navigation.
    Experiment [c] explores collaborative experience with 1:1 weight ratio between $\xi^{PC}$ and $\xi^{OVC}$, and experiment [d] only explores a unitary feedback loop $r = -1.0$.
    It is left for further work to explore network architecture theory or find data-driven methods for parameter adaptation.
    We have barely scratched the surface of autonomous navigation by purposive networks, proposing neoRL networks as a plausible new approach toward navigational autonomy.

\footnotesize
\bibliographystyle{plain}
\bibliography{bibliography}
\end{document}